\documentclass[11pt]{article}
\usepackage{naaclhlt2013}
\usepackage{times}
\usepackage{latexsym}
\usepackage{mathtools}  
\usepackage{url} 
\title{Sentiment Analysis: How to Derive Prior Polarities from SentiWordNet    
    }

\author{Marco Guerini\\
	    Trento RISE\\
	    Via Sommarive 18\\
	    38123 Povo in Trento, Italy\\
	   {\small  {\tt m.guerini@trentorise.eu}}
	  \And
	  Lorenzo Gatti\\
	    Trento RISE\\
	    Via Sommarive 18\\
	    38123 Povo in Trento, Italy\\
	   {\small  {\tt l.gatti@trentorise.eu}}
	  \And
	Marco Turchi\\
  	Fondazione Bruno Kessler\\
	    Via Sommarive 18\\
	    38123 Povo in Trento, Italy\\
  {\small {\tt turchi@fbk.eu}}}

\date{}

\begin{document}
\maketitle
\begin{abstract}
Assigning a positive or negative score to a word out of context (i.e. a word's prior polarity) is a challenging task for sentiment analysis. In the literature, various approaches based on SentiWordNet have been proposed. 
In this paper, we compare the most often used techniques together with newly proposed ones and incorporate all of them in a learning framework to see whether blending them can further improve the estimation of prior polarity scores.  Using two different versions of SentiWordNet and testing regression and classification models across tasks and datasets, our learning approach consistently outperforms the single metrics, providing a new state-of-the-art approach in computing words' prior polarity for sentiment analysis. We conclude our investigation showing interesting biases in calculated prior polarity scores when word Part of Speech and annotator gender are considered. 
\end{abstract}

\section{Introduction}

Many approaches to sentiment analysis make use of lexical resources -- i.e. lists of positive and negative words -- often deployed as baselines or as features for other methods  (usually machine learning based) for sentiment analysis research  \cite{liu2012survey}. In these lexica, words are associated with their prior polarity, i.e. if that word out of context evokes something positive or something negative. For example, \emph{wonderful} has a positive connotation -- prior polarity -- while \emph{horrible} has a negative one. These approaches have the advantage of not needing deep semantic analysis or word sense disambiguation to assign an affective score to a word and are domain independent (they are thus less precise but more portable).

SentiWordNet (henceforth SWN) is one of these resources and has been widely adopted since it provides a broad-coverage lexicon -- built in a semi-automatic manner -- for English \cite{Esuli06}. 
Given that SWN provides polarities scores for each word sense (also called `posterior polarities'), it is necessary to derive prior polarities from the posteriors. For example, the word \emph{cold} has a posterior polarity for the meaning ``having a low temperature'' -- like in ``\emph{cold} beer'' -- that is different from the one in ``\emph{cold} person'' which refers to ``being emotionless''. This information must be considered when reconstructing the prior polarity of \emph{cold}.

Several formulae to compute prior polarities starting from posterior polarities scores have been used in the literature. However, their performance varies significantly depending on the adopted variant. We show that researchers have not paid sufficient attention to this \emph{posterior-to-prior polarity} issue. Indeed, we show that  some variants outperform others on different datasets and can represent a fairer state-of-the-art approach using SWN. On top of this, we attempt to outperform 
the state-of-the-art formula using a learning framework that combines the various formulae together.

In detail, we will address five main research questions:  
(\emph{i}) is there any relevant difference in the posterior-to-prior polarity formulae performance (both in regression and classification tasks),
(\emph{ii}) is there any relevant variation in prior polarity values if we use different releases of SWN (i.e. $SWN_1$ or $SWN_3$), 
(\emph{iii}) can a learning framework boost performance of such formulae,
(\emph{iv}) considering word Part of Speech (PoS), is there any relevant difference in formulae performance,
(\emph{v}) considering the gender dimension of the annotators (male/female) and the sentiment dimension (positive/negative), is there any relevant difference in SWN performance.

In Section \ref{sec:approach} we briefly describe our approach and how it differentiates from similar sentiment analysis tasks. Then, in Sections \ref{sec:SWN}  and \ref{sec:PriorF}, we present SentiWordNet 
 and overview various posterior-to-prior polarity formulae based on this resource that appeared in the literature (included some new ones we identified as potentially relevant). In Section \ref{sec:ML} we describe the learning approach adopted on prior-polarity formulae.  In Section \ref{sec:GOLD} we introduce the ANEW  and General Inquirer resources that will be used as gold standards. Finally, in the two last sections, we present a series of experiments, both in regression and classification tasks, that give an answer to the aforementioned research questions. The results support the hypothesis that using a learning framework we can improve on state-of-the-art performance and that there are some interesting phenomena connected to PoS and annotator gender.

\section{Proposed Approach}
\label{sec:approach}

In the broad field of Sentiment Analysis we will focus on the specific problem of posterior-to-prior polarity assessment, using both regression and classification experiments. A general overview on the field and possible approaches can be found in \cite{pang2008opinion} or  \cite{liu2012survey}.

For the regression task, we tackled the problem of assigning affective scores (along a continuum between -1 and 1) to words using the posterior-to-prior polarity formulae. For the classification task (assessing whether a word is either \emph{positive} or \emph{negative}) we used the same formulae, but considering just the sign of the result.  In these experiments we will also use a learning framework which combines the various formulae together. The underlying hypothesis is that by blending these formulae, and looking at the same information from different perspectives (i.e. the posterior polarities provided by SWN combined in various ways), we can give a better prediction.

The regression task is harder than binary classification,  
since we want to assess not only that \emph{pretty}, \emph{beautiful} and \emph{gorgeous} are positive words, but also to define a partial or total order so that \emph{gorgeous} is more positive than \emph{beautiful} which, in turn, is more positive than \emph{pretty}. This is fundamental for tasks such as affective modification of existing texts, where words' polarity together with their score are necessary for creating multiple graded variations of the original text \cite{Guerini2008}. Some of the work that addresses the problem of sentiment strength are presented in \cite{wilson:AAAI-04,Paltoglou2010}, however, their approach is modeled as a multi-class classification problem (\emph{neutral}, \emph{low}, \emph{medium} or \emph{high} sentiment) at the sentence level, rather than a regression problem at the word level. Other works such as \cite{neviarouskaya2011affect} use a fine grained classification approach too, but they consider emotion categories (\emph{anger}, \emph{joy}, \emph{fear}, etc.), rather than sentiment strength categories. 
On the other hand, even if approaches that go beyond pure prior polarities -- e.g. using word bigram features \cite{wangbaselines} -- are better for sentiment analysis tasks, there are tasks that are intrinsically based on the notion of words' prior polarity. Consider copywriting, where evocative names are a key element to a successful product \cite{ozbalcomputational,ozbal2012brand}. In such cases no context is given and the brand name alone, with its perceived prior polarity, is responsible for stating the area of competition and evoking semantic associations. For example \emph{Mitsubishi} changed the name of one of its SUV for the Spanish market, since the original name \emph{Pajero} had a very negative prior polarity, as it meant `wanker' in Spanish \cite{piller200310}.

To our knowledge, the only work trying to address the SWN
posterior-to-prior polarity issue, comparing some of the approaches
appeared in the literature is \cite{gatti-guerini:2012:POSTERS}.
However, in our previous study we only considered a regression
framework, we did not use machine learning and we only tested $SWN_1$.
So, we took this work as a starting point for our analysis and expanded on it.

	
\section{SentiWordNet}
\label{sec:SWN}


SentiWordNet \cite{Esuli06} is a lexical resource in which each entry is a set of lemma-PoS pairs sharing the same meaning, called ``synset". Each synset \texttt{s} is associated with the numerical scores \texttt{Pos(s)} and \texttt{Neg(s)}, which range from 0 to 1.  These scores -- automatically assigned starting from a bunch of seed terms -- represent the positive and negative valence (or posterior  polarity) of the synset and are inherited by each lemma-PoS in the synset. According to the structure of SentiWordNet, each pair can have more than one sense and each of them takes the form of \texttt{lemma\#PoS\#sense-number}, where the smallest sense-number corresponds to the most frequent sense.  

%
%

Obviously, different senses can have different polarities. In Table \ref{tab:sentiwncoldsenses}, the first 5 senses of \texttt{cold\#a} present all  possible combinations, included mixed scores (\texttt{cold\#a\#4}), where positive and negative valences are assigned to the same sense. Intuitively, mixed scores for the same sense are acceptable, as in ``\emph{cold} beer'' (positive) vs. ``\emph{cold} pizza'' (negative).

 
\begin{table} [h]

	\centering
	{\scriptsize
		\begin{tabular}{ccccl}
      \hline\hline
      \textbf{PoS}&\textbf{Offset}&\textbf{Pos(s)}&\textbf{Neg(s)}&\textbf{SynsetTerms}\\
      \hline\hline
      a&1207406&0.0&0.75&\texttt{cold\#a\#1}\\
      a&1212558&0.0&0.75&\texttt{cold\#a\#2}\\
      a&1024433&0.0&0.0&\texttt{cold\#a\#3}\\
      a&2443231&0.125&0.375&\texttt{cold\#a\#4}\\
      a&1695706&0.625&0.0&\texttt{cold\#a\#5}\\
      \hline
		\end{tabular}
		}
	
	\setlength{\belowcaptionskip}{-0.1cm}
	\caption{First five \textit{SentiWordNet} entries for \texttt{cold\#a}}
	\label{tab:sentiwncoldsenses}
\end{table}

In our experiments we use two different versions of SWN: SentiWordNet 1.0 ($SWN_1$), the first release of SWN, and its updated version SentiWordNet 3.0 \cite{baccianella2010sentiwordnet} --  $SWN_3$. In $SWN_3$ the annotation algorithm used in $SWN_1$ was revised, leading to an increase in the accuracy of posterior polarities over the previous version.

\section{Prior Polarities Formulae}
\label{sec:PriorF}


In this section we review the main strategies for computing prior polarities used in previous studies. All the proposed approaches try to estimate the prior polarity score from the posterior polarities of all the senses for a single lemma-PoS. Given a lemma-PoS with $n$ senses (\texttt{lemma\#PoS\#n}), every formula $f$ is independently applied to all the \texttt{Pos(s)} and \texttt{Neg(s)} . This produces two scores, $f(posScore)$ and $f(negScore)$, for each lemma-PoS. To obtain a unique prior polarity for each lemma-PoS, $f(posScore)$ and $f(negScore)$ can be mapped according to different strategies:
\begin{align*}
f_m &=
\begin{cases}
\hphantom{-}f(posScore) & \text{if  $f(posScore)\geq$}\\
& \text{$f(negScore)$}\\
-f(negScore) & \text{otherwise}
\end{cases}\\[5pt]
f_d &= f(posScore) - f(negScore)
\end{align*}
  where $f_m$ computes the absolute maximum of the two scores, while $f_d$ computes the difference between them. It is worth noting that  $f(negScore)$ is always positive by construction. To obtain a final prior polarity that ranges from -1 to 1, the negative sign is imposed.  So, considering the first 5 senses of \texttt{cold\#a} in Table \ref{tab:sentiwncoldsenses}, $f(posScore)$ will be derived from the \texttt{Pos(s)}  values {\small ${<}0.0 , 0.0 , 0.0 , 0.125 , 0.625{>}$}, while $f(negScore)$ from {\small ${<}0.750 , 0.750 , 0.0 , 0.375 , 0.0{>}$}. Then, the final polarity strength returned will be either $f_m$ or $f_d$.\\

The formulae ($f$) we tested are the following:

\textbf{fs.} In this formula only the first (and thus most frequent) sense is considered for the given \texttt{lemma\#PoS}. This is equivalent to considering only the SWN score for \texttt{lemma\#PoS\#1}. Based on \cite{neviarouskaya2009sentiful,agrawal2009using,Guerini2008,chowdhury-EtAl:2013:SemEval-2013}, 
this is the most basic form of prior polarities. 



\textbf{mean.} It calculates the mean of the positive and negative scores for all the senses of the given \texttt{lemma\#PoS}. This formula has been used in  \cite{thet2009sentiment,denecke2009sentiwordnet,devitt2007sentiment,sing2012development}. 


\textbf{uni.} Based on \cite{neviarouskaya2009sentiful}, it considers only those senses that have a \texttt{Pos(s)} greater than or equal to the corresponding \texttt{Neg(s)}, and greater than 0 (the $stronglyPos$ set). In case $posScore$ is equal to $negScore$, the one with the highest weight is returned, where weights are defined as the cardinality of $stronglyPos$ divided by the total number of senses. The same applies for the negative senses. This is the only method, together with $rnd$, for which we cannot apply $f_d$, as it returns a positive or negative score according to the weight.

\textbf{uniw.} Like $uni$ but without the weighting system.

\textbf{w1.} This formula weighs each sense with a geometric series of ratio 1/2. The rationale behind this choice is based on the assumption that more frequent senses should bear more ``affective weight'' than rare senses when computing the prior polarity of a word.  The system presented in \cite{chaumartin2007upar7} uses a similar approach of weighted mean.

\textbf{w2.} Similar to the previous one, this formula weighs each lemma with a harmonic series, see for example \cite{denecke2008accessing}.

\medskip
On top of these formulae, we implemented some new formulae that were relevant to our task and have not been implemented before. 
These formulae mimic the ones discussed previously, but they are built under a different assumption: that the saliency \cite{giora1997understanding} of a word's prior polarity might be more related to its posterior polarities score, rather than to sense frequencies.
Thus we ordered $posScore$ and $negScore$ by strength, giving more relevance to `valenced' senses.
For instance, in Table \ref{tab:sentiwncoldsenses}, $posScore$ and $negScore$ for \texttt{cold\#a} become {\small ${<}0.625 , 0.125 , 0.0 , 0.0 , 0.0{>}$} and {\small ${<}0.750 , 0.750 , 0.375 , 0.0 , 0.0{>}$} respectively.

\textbf{w1s} and \textbf{w1n}. Like $w1$ and $w2$, but senses are ordered by strength (sorting {Pos(s)} and \texttt{Neg(s)} independently).

\textbf{w1n} and \textbf{w2n}. Like $w1$ and $w2$ respectively, but without considering senses that have a 0 score for both \texttt{Pos(s)} and \texttt{Neg(s)}. Our motivation is that ``empty" senses are mostly noise.

\textbf{w1sn} and \textbf{w2sn}. Like $w1s$ and $w2s$, but without considering senses that have a 0 score for both \texttt{Pos(s)} and \texttt{Neg(s)}.

\textbf{median}: return the median of the senses ordered by polarity score.

\medskip

All these prior polarities formulae are compared against two gold
standards (one for regression, one for classification) both one by
one, as in the works mentioned above,  and combined together in a
learning framework (to see whether combining these features -- that
capture different aspect of prior polarities --  can further improve
the results).

\medskip

Finally, we implemented two variants of a prior polarity random baseline to asses
possible advantages of approaches using SWN:

\textbf{rnd.} This formula represents the basic baseline random
approach. It  simply returns a random number between -1 and 1 for any
given \texttt{lemma\#PoS}.

\textbf{swnrnd.} This formula represents an advanced random approach
that incorporates some ``knowledge'' from SWN. It takes the scores of a random sense for the given \texttt{lemma\#PoS}. We believe this is a fairer baseline than $rnd$ since SWN information can possibly constrain the values. A similar approach has been used in  \cite{qu2008sentence}.

\section{Learning Algorithms}
\label{sec:ML}

We used two non-parametric learning approaches, Support Vector Machines (SVMs) \cite{shawe2004kernel} and Gaussian Processes (GPs) \cite{2006gaussian}, to test the performance of all the metrics in conjunction.  SVMs are non-parametric deterministic algorithms that have been widely used in several fields, in particular in NLP where they are the state-of-the-art for various tasks. GPs, on the other hand, are an extremely flexible non-parametric probabilistic framework able to explicitly model uncertainty, that,  despite being considered state-of-the-art in regression,  have rarely been used in NLP. To our knowledge only two previous works did so \cite{polajnar2011protein,cohnmodelling}. 

Both methods take advantage of the \textit{kernel trick}, a technique used to embed the original feature space into an alternative space where data may be linearly separable. This is performed by the kernel function that transforms the input data in a new structure, called \textit{kernel}. How it is used to produce  the prediction is one of the main differences between SVMs and GPs. In classification SVMs use the geometric mean to discriminate between the positive and negative classes, while the GP model uses the posterior probability distribution over each class. Both frameworks support learning algorithms for regression and classification. 
An exhaustive explanation of the two methodologies can be found in \cite{shawe2004kernel}  and \cite{2006gaussian}.

In the SVM experiments, we use $C$-SVM and $\epsilon$-SVM implemented in the LIBSVM toolbox \cite{CC01a}. The selection of the kernel (linear, polynomial, radial basis function and sigmoid)  and the optimization of the parameters are carried out through grid search in 10-fold cross-validation.

GP regression models with Gaussian noise are a rare exception where the exact inference with likelihood functions is tractable, see \S2 in \cite{2006gaussian}. Unfortunately, this is not valid for the classification task -- see \S3 in \cite{2006gaussian} -- where an approximation method is required. In this work, we use the Laplace approximation method proposed in \cite{williams1998bayesian}. Different kernels are tested (covariance for constant functions, linear with and without automatic relevance determination (ARD)\footnote{$linone$ and $linard$ in the result tables, respectively.}, Matern, neural network, etc.\footnote{More detailed information on the available kernels are in $\S$4 \cite{2006gaussian}})
and the linear logistic ($lll$) and probit regression ($prl$) likelihood functions are evaluated in classification. In our classification experiments we tried all possible combinations of kernels and likelihood functions, while in the regression tests we ranged only on different kernels.  All the GP models were implemented using the GPML Matlab toolbox.
Unlike SVMs, the optimization of the kernel parameters can be performed without using grid search, but the optimal parameters can be obtained iteratively, by maximizing the marginal likelihood (or in classification, the Laplace approximation of the marginal likelihood).  We fix at 100 the maximum number of iterations. 

An interesting property of the GPs is their capability of weighting the features differently according to their importance in the data. This is referred to as the automatic variance determination kernel. As demonstrated in \cite{weston2000feature}, SVMs can benefit from the application of feature selection techniques especially when there are highly redundant features. Since the prior polarities formulae tend to cluster in groups that provide similar results  \cite{gatti-guerini:2012:POSTERS}  -- creating noise for the learner -- we
want to understand whether feature selection approaches can boost the performance of SVMs. For this reason, 
we also test feature selection prior to the SVM  training. For that we used Randomized Lasso, or stability selection \cite{Meinshausen2010}. Re-sampling of the training data is performed several times and a Lasso regression model is fit on each sample.  Features that appear in a given number of samples are retained. Both the fraction of the data to be sampled and the threshold to select the features can be configured. In our experiments we set the sampling fraction to 75\%, the selection threshold to 25\% and the number of re-samples to 1,000. We refer to 
these as \textit{SVMfs}. 


\section{Gold Standards}
\label{sec:GOLD}

To assess how well prior polarity formulae perform, a gold standard with word polarities provided by human annotators is
needed. There are
many such resources in the literature, each with different coverage
and annotation characteristics. ANEW \cite{bradley1999affective} rates the
valence score of 1,034 words, which were presented in isolation to
annotators. The SO-CAL entries \cite{taboada2011lexicon} were
collected from corpus data and then manually tagged by a small number
of annotators with a multi-class label. These ratings were further
validated through crowdsourcing. Other resources, such as the General
Inquirer lexicon \cite{stone1966general}, provide a binomial
classification  (either \emph{positive} or \emph{negative}) of
sentiment-bearing words. The resource presented in
\cite{wilson2005recognizing} uses a similar binomial annotation for
single words; another interesting resource is WordNetAffect
\cite{strappaLREC04} 
but it labels words senses and it cannot be used for the prior polarity validation task. 

In the following we describe in detail the two resources we used for our experiments, namely ANEW for the regression experiments and the General Inquirer (GI) for the classification ones.

\subsection{ANEW}
\label{sec:AN}

ANEW \cite{bradley1999affective} is a resource developed to provide a set of normative emotional ratings for a large number of words (roughly 1 thousand) in the English language. It contains a set of words that have been rated in terms of pleasure (affective valence), arousal, and dominance. In particular for our task we considered the valence dimension. Since words were presented to subjects in isolation (i.e. no context was provided) this resource represents a human validation of prior polarities scores for the given words, and can be used as a gold standard. For each word ANEW provides two main metrics: $anew_\mu$, which correspond to the average of annotators votes, and $anew_\sigma$, which gives the variance in annotators scores for the given word. In the same way these metrics are also provided for the male/female annotator groups. 

\subsection{General Inquirer}
\label{sec:GI}

The Harvard General Inquirer dictionary 
is a widely used resource, built for automatic text analysis
\cite{stone1966general}. Its latest
revision\footnote{\url{http://www.wjh.harvard.edu/~inquirer/}}
contains 11789 words, tagged with 182 semantic and pragmatic labels,
as well as with their part of speech. Words and their categories were
initially taken from the Harvard IV-4 Psychosociological Dictionary
\cite{harvardiv} and the Lasswell Value Dictionary
\cite{lasswell1969lasswell}.
For this paper we consider the \texttt{Positiv} and \texttt{Negativ} categories (1,915 words the former, 2,291 words the latter,
for a total of 4,206 affective words).

\section{Experiments}
\label{sec:exp}

In order to use the ANEW dataset to measure prior polarities formulae
performance, we had to assign a PoS to all the words to obtain
 the SWN \texttt{lemma\#PoS} format. To do
so, we proceeded as follows: for each word, check if it is present
among both $SWN_1$ and $SWN_3$ lemmas; if not,
lemmatize the word with the TextPro tool suite
\cite{pianta2008textpro} and check if the lemma is present
instead\footnote{We did not lemmatize everything to avoid duplications
(for example, if we lemmatize the ANEW entry \emph{addicted}, we obtain
\emph{addict}, which is already present in ANEW).}. If it is not found
(i.e., the word cannot be aligned automatically), remove the word from
the list (this was the case for 30 words of the 1,034 present in
ANEW). The remaining 1,004 lemmas were then associated with all the PoS
present in SWN to get the final \texttt{lemma\#PoS}. Note
that a lemma can have more than one PoS,  for example, \emph{writer} is
present only as a noun (\texttt{writer\#n}), while \emph{yellow} is present
as a verb, a noun and an adjective (\texttt{yellow\#v},
\texttt{yellow\#n}, \texttt{yellow\#a}). This gave us a list of 1,484
words in the \texttt{lemma\#PoS} format. 

In a similar way we
pre-processed the GI words that 
uses the generic \texttt{modif} label to indicate either adjective or adverb
(\texttt{noun}  and \texttt{verb}  PoS were instead consistently
used).
Finally, all the sense-disambiguated words in the \texttt{lemma\#PoS\#n} format were discarded (1,114
words out of the 4,206 words with positive or negative valence). 

After the two datasets were built this way, we removed the words for
which the $posScore$ and $negScore$ contained all 0 in both $SWN_1$
and $SWN_3$ (523 \texttt{lemma\#PoS} for ANEW and  484 for the
GI dataset), since these words are not informative for our experiments. The final dataset included 961 entries for ANEW and
2,557 for GI. For each \texttt{lemma\#PoS} in GI and ANEW, we then applied the prior polarity formulae
described in Section \ref{sec:PriorF}, using both $SWN_1$ and $SWN_3$
and annotated the results.


According to the nature of the human labels (real numbers or -1/1), we ran several regression and classification experiments. In both cases, each dataset was randomly split into 70\% for training and the remaining for test. This process was repeated 5 times to generate different splits. For each partition, optimization of the learning algorithm parameters was performed on the training data (in 10-fold cross-validation for SVMs). Training and test sets were normalized using the z-score.

To evaluate the performance of our regression experiments on ANEW we used the Mean Absolute Error ($MAE$), that averages the error over a given test set. 
Accuracy was used for the classification experiments on GI instead. We opted for accuracy -- rather than F1 -- since for us True Negatives have same importance as True Positives. For each experiments we reported the average performance and the standard deviation over the 5 random splits. In the following sections, to check if there was a statistically significant difference in the results, we used Student's t-test for regression experiments, while an approximate randomization test \cite{yeh2000more} was used for the  classification experiments.

In Tables \ref{tab:ANEW_$SWN_1$} and \ref{tab:ANEW_$SWN_3$}, the results of regression experiments over the ANEW dataset, using $SWN_1$ and $SWN_3$, are presented. The results of the classification experiments over the GI dataset, using $SWN_1$ and $SWN_3$ are shown in Tables \ref{tab:GI_$SWN_1$} and \ref{tab:GI_$SWN_3$}. For the sake of interpretability, results are divided according to the main approaches: randoms, posterior-to-prior formulae, learning algorithms.  Note that for classification we report the generics $f$ and not the $f_m$ and $f_d$ variants. In fact, both versions always return the same classification answer (we are classifying according to the sign of $f$ result and not its strength). For the GPs, we report the two best configurations only.

\begin{table} [ht!]

	\centering
	{\footnotesize
		\begin{tabular}{l|rr}
      \hline\hline
  & MAE $\mu$ & MAE $\sigma$ \\
 \hline
 $rnd$ & 0.652 & 0.026\\ 
$swnrnd_m$ & 0.427 & 0.011 \\
$swnrnd_d$ & 0.426 & 0.009 \\
 \hline
$uniw_m$ & 0.420 & 0.009 \\
$max_m$ & 0.419 & 0.009 \\
$fs_d$ & 0.413 & 0.011 \\
$fs_m$ & 0.412 & 0.009 \\
$uni$ & 0.410 & 0.010 \\
$uniw_d$ & 0.406 & 0.007 \\
$w1sn_m$ & 0.405 & 0.011 \\
$max_d$ & 0.404 & 0.005 \\
$w2sn_m$ & 0.402 & 0.011 \\
$median_d$ & 0.401 & 0.014 \\
$w1_d$ & 0.401 & 0.010 \\
$w1n_d$ & 0.399 & 0.008 \\
$mean_d$ & 0.398 & 0.010 \\
$w2_d$ & 0.398 & 0.010 \\
$median_m$ & 0.397 & 0.015 \\
$w1sn_d$ & 0.397 & 0.008 \\
$w2sn_d$ & 0.397 & 0.008 \\
$w2n_d$ & 0.397 & 0.008 \\
$w1s_m$ & 0.396 & 0.010 \\
$w1_m$ & 0.396 & 0.010 \\
$w1n_m$ & 0.394 & 0.009 \\
$mean_m$ & 0.393 & 0.011 \\
$w2s_d$ & 0.393 & 0.008 \\
$w1s_d$ & 0.393 & 0.009 \\
$w2s_m$ & 0.392 & 0.010 \\
$w2_m$ & 0.391 & 0.011 \\
$w2n_m$ & 0.391 & 0.012 \\
\hline
$GP_{linard}$ & 0.398 & 0.014 \\
$GP_{linone}$ & 0.398 & 0.014 \\
$SVM$ & 0.367 & 0.010 \\
$SVMfs$ & 0.366 & 0.011 \\
 \hline
AVERAGE & 0.398 & 0.010\\
 \hline
		\end{tabular}
		}
	
	\setlength{\belowcaptionskip}{-0.1cm}
	\caption{MAE results for metrics using $SWN_1$}
	\label{tab:ANEW_$SWN_1$}
\end{table}

\begin{table} [h]

	\centering
	{\footnotesize
		\begin{tabular}{l|rr}
      \hline\hline
  & MAE $\mu$ & MAE $\sigma$ \\
 \hline
$rnd$ & 0.652 & 0.026\\ 
$swnrnd_d$ & 0.404 & 0.013 \\
$swnrnd_m$ & 0.402 & 0.010 \\
\hline
$max_m$ & 0.393 & 0.009 \\
$fs_d$ & 0.382 & 0.008 \\
$uniw_m$ & 0.382 & 0.015 \\
$fs_m$ & 0.381 & 0.010 \\
$median_m$ & 0.377 & 0.008 \\
$uniw_d$ & 0.377 & 0.012 \\
$median_d$ & 0.377 & 0.011 \\
$uni$ & 0.376 & 0.010 \\
$max_d$ & 0.372 & 0.011 \\
$mean_d$ & 0.371 & 0.010 \\
$w1sn_m$ & 0.371 & 0.011 \\
$w2sn_m$ & 0.369 & 0.010 \\
$w1_d$ & 0.368 & 0.010 \\
$w2_d$ & 0.367 & 0.010 \\
$mean_m$ & 0.367 & 0.010 \\
$w1_m$ & 0.365 & 0.010 \\
$w2sn_d$ & 0.364 & 0.011 \\
$w1sn_d$ & 0.364 & 0.010 \\
$w1s_m$ & 0.363 & 0.009 \\
$w1n_d$ & 0.362 & 0.009 \\
$w2s_d$ & 0.362 & 0.010 \\
$w2_m$ & 0.362 & 0.010 \\
$w1s_d$ & 0.362 & 0.009 \\
$w1n_m$ & 0.362 & 0.007 \\
$w2n_d$ & 0.361 & 0.010 \\
$w2s_m$ & 0.360 & 0.009 \\
$w2n_m$ & 0.359 & 0.009 \\
\hline
$GP_{linone}$ & 0.356 & 0.008 \\
$GP_{linard}$ & 0.355 & 0.008 \\
$SVM$ & 0.333 & 0.004 \\
$SVMfs$ & 0.333 & 0.003 \\
 \hline
AVERAGE & 0.366 & 0.009 \\
 \hline
		\end{tabular}
		}
	
	\setlength{\belowcaptionskip}{-0.1cm}
	\caption{MAE results for regression using $SWN_3$}
	\label{tab:ANEW_$SWN_3$}
\end{table}

\begin{table} [h]

	\centering
	{\footnotesize
		\begin{tabular}{l|rr}
      \hline\hline
 & Acc. $\mu$ & Acc. $\sigma$ \\ 
\hline
$rnd$ & 0.447 & 0.019\\ 
$swn\_rnd_m$ & 0.639 & 0.026 \\
$swn\_rnd_d$ & 0.646 & 0.021 \\
\hline
$fs\_m$ & 0.659 & 0.020 \\
$uni$ & 0.684 & 0.017 \\
$median$ & 0.686 & 0.022 \\
$uniw$ & 0.702 & 0.019 \\
$max$ & 0.710 & 0.022 \\
$w1$ & 0.712 & 0.021 \\
$w1n$ & 0.713 & 0.022 \\
$w2n$ & 0.714 & 0.023 \\
$w2$ & 0.715 & 0.021 \\
$mean$ & 0.718 & 0.023 \\
$w2s$ & 0.719 & 0.023 \\
$w2sn$ & 0.719 & 0.023 \\
$w1s$ & 0.719 & 0.023 \\
$w1sn$ & 0.719 & 0.023 \\
\hline
$GP_{linard}^{lll}$ & 0.721 & 0.026 \\
$GP_{linard}^{prl}$ & 0.722 & 0.025 \\
$SVM$ & 0.733 & 0.021 \\
$SVMfs$ & 0.743 & 0.021 \\
\hline
Average & 0.710 & 0.022\\ 
\hline
		\end{tabular}
		}
	
	\setlength{\belowcaptionskip}{-0.1cm}
	\caption{Accuracy results for classification using $SWN_1$}
	\label{tab:GI_$SWN_1$}
\end{table}

\begin{table} [h]

	\centering
	{\footnotesize
		\begin{tabular}{l|rr}
      \hline\hline
& Acc. $\mu$ &	Acc. $\sigma$ \\ 
 \hline
 $rnd$ & 0.447 & 0.019\\ 
 $swn\_rnd_d$ & 0.700 & 0.030 \\
 $swn\_rnd_m$ & 0.706 & 0.034 \\
\hline
$fs$ & 0.723 & 0.014 \\
$medianm$ & 0.742 & 0.016 \\
$uni$ & 0.750 & 0.015 \\
$uniw$ & 0.762 & 0.023 \\
$max$ & 0.769 & 0.019 \\
$w2s$ & 0.777 & 0.017 \\
$w2sn$ & 0.777 & 0.017 \\
$w1s$ & 0.777 & 0.017 \\
$w1sn$ & 0.777 & 0.017 \\
$w1n$ & 0.780 & 0.021 \\
$w2n$ & 0.780 & 0.022 \\
$mean$ & 0.781 & 0.018 \\
$w1$ & 0.781 & 0.021 \\
$w2$ & 0.781 & 0.021 \\
\hline
$SVM$ & 0.779 & 0.016 \\
$GPl$ & 0.779 & 0.018 \\
$GPg$ & 0.781 & 0.018 \\
$SVMfs$ & 0.792 & 0.014 \\
\hline
 Average & 0.771 & 0.018\\ 
 \hline
		\end{tabular}
		}
	
	\setlength{\belowcaptionskip}{-0.1cm}
	\caption{Accuracy results for classification using $SWN_3$}
	\label{tab:GI_$SWN_3$}
\end{table}

\section{General Discussion} 
\label{sec:class_exp}

In this section we sum up the main results of our analysis, providing an answer to the various questions we introduced at the beginning of the paper:

\textbf{SentiWordNet improves over random.} One of the first things worth noting -- in Tables \ref{tab:ANEW_$SWN_1$}, \ref{tab:ANEW_$SWN_3$}, \ref{tab:GI_$SWN_1$} and \ref{tab:GI_$SWN_3$} -- is that the random approach (\emph{rnd}), as expected, is the worst performing metric, while all other approaches, based on \emph{SWN}, have statistically significant improvements both for MAE and for Accuracy ($p<0.001$). So, using \emph{SWN} for posterior-to-prior polarity computation brings benefits, since it increases the performance above the baseline in words' prior polarity assessment.

\textbf{SWN$_3$ is better than SWN$_1$.} With respect to $SWN_1$, using $SWN_3$ enhances performance, both in regression (MAE $\mu$ 0.398 vs. 0.366, $p < 0.001$) and classification (Accuracy $\mu$ 0.710 vs. 0.771, $p < 0.001$) tasks. Since many of the approaches described in the literature use $SWN_1$ their results should be revised and $SWN_3$ should be used as standard. This difference in performance can be partially explained by the fact that, even after pre-processing, for the ANEW dataset 137 \texttt{lemma\#PoS} have all senses equal to 0 in $SWN_1$, while in $SWN_3$ they are just 48. In the GI lexicon the numbers are 233 for $SWN_1$ and 69 for $SWN_3$.

\textbf{Not all formulae are created equal.} The formulae described in Section \ref{sec:PriorF} have very different results, along a continuum. While inspecting every difference in performance is out of the scope of the present paper, we can see that there is a strong difference between best and worst performing formulae both in regression (in Table \ref{tab:ANEW_$SWN_1$} $w2n_m$ is better than $uniw_m$, 
in Table \ref{tab:ANEW_$SWN_3$} $w2n_m$ is better than $max_m$) 
and classification (in Table \ref{tab:GI_$SWN_1$} $w1sn_m$ is better than $fs_m$,
in Table \ref{tab:GI_$SWN_3$} $w2_m$ is better than $fs_m$
) and these differences are all statistically significant ($p < 0.001$). Again, these results indicate that the previous experiments in the literature that use SWN as a baseline should be revised to take these results into account. Furthermore, the new formulae we introduced, based on the ``posterior polarities saliency'' hypothesis,  proved to be among the best performing in all experiments. This entails that there is room for inspecting new formulae variants other than those already proposed in the literature. 



\textbf{Selecting just one sense is not a good choice.} On a side note, the approaches that rely on only one sense polarity (namely $fs$, $median$ and $max$) have similar results which do not differ significantly from $swnrnd$ (for $max_m$, $fs_d$ and $fs_m$ in Table \ref{tab:ANEW_$SWN_1$}, and for $max_m$ in Table \ref{tab:ANEW_$SWN_3$}). These same approaches are also far from the best performing formulae: in Table \ref{tab:ANEW_$SWN_3$}, $median_d$ differs from $w2n_m$ ($p < 0.05$), as do $max_m$, $max_d$, $fs_m$ and $fs_d$ ($p < 0.001$); in Table \ref{tab:ANEW_$SWN_3$}, $fs$, $max$ and $median$ in both their $f_m$ and $f_d$ variants are significantly different from the best performing $w2n_m$ ($p < 0.001$). For classification, in Table \ref{tab:GI_$SWN_1$} and \ref{tab:GI_$SWN_3$} the difference between the corresponding best performing formula and the single senses formulae is always significant (at least $p < 0.01$). 
Among other things, this finding entails, surprisingly, that taking the first sense of a \texttt{lemma\#PoS} in some cases has no improvement over taking a random sense, and that in all cases it is one of the worst approaches with $SWN$. This is surprising since in many NLP tasks, such as word sense disambiguation, algorithms based on most frequent sense represent a very strong baseline\footnote{In SemEval 2010, only 5 participants out of 29 performed better than the most frequent threshold \cite{agirre2010semeval}.}.  

\textbf{Learning improvements.} Combining the formulae in a learning framework further improves the results over the best performing formulae, both in regression (MAE$\mu$ with $SWN_1$ 0.366 vs. 0.391, $p < 0.001$; MAE$\mu$ with $SWN_3$ 0.333 vs. 0.359, $p < 0.001$) and in classification (Accuracy$\mu$ for $SWN_1$ is 0.743 vs. 0.719, $p < 0.001$; Accuracy$\mu$ for $SWN_3$ is 0.792 vs. 0.781, not significant $p = 0.07$). 
Another thing worth noting is that, in regression, GPs are outperformed by both versions of SVM ($p < 0.001$), see Tables \ref{tab:ANEW_$SWN_1$} and \ref{tab:ANEW_$SWN_3$}.  This is in contrast with the results presented in \cite{cohnmodelling}, where GPs on the single task are on average better than SVMs. In classification, GPs have similar performance to SVM without feature selection, and in some cases (see Table \ref{tab:GI_$SWN_3$}) even slightly better. 
Analyzing the selected kernels for GPs and SVMs, we notice that in most of the splits SVMs prefer the radial based function, while the best performance with the GPs are obtained with linear kernels with and without ARD. There is no significant difference in using linear logistic and probit regression likelihoods.
In all our experiments, SVM with feature selection leads to the best performance. This is not surprising due the high level of redundancy in the formulae scores. Interestingly, inspecting the most frequent selected features by $SVMfs$, we see that features from different groups are selected, and even the worst performing formulae can add information, 
confirming the idea that viewing the same information from different perspectives (i.e. the posterior polarities provided by SWN combined in various ways) can give better predictions. 

\medskip
 
To sum up: the new state-of-the-art performance level in prior-polarity computation is represented by the $SVMfs$ approach using $SWN_3$, and this should be used as the reference from now on.

\section{PoS and Gender Experiments} 
\label{sec:2_exp}

Next, we wanted to understand if the performance
of our approach, using $SWN_3$, was consistent across 
word PoS. In Table \ref{tab:GI_POS} we report the
results for the best performing formulae and learning algorithm on the
GI PoS classes. In particular for ADJ there are 1,073 words, 922
for NOUN  and 508 for VERB. We discarded adverbs since the class
was too small to allow reliable evaluation and efficient learning
(only 54 instances). The results show a greater accuracy for adjectives ($p<0.01$), while performance for nouns and verbs are
similar. 

\begin{table} [h]

	\centering
	{\footnotesize
		\begin{tabular}{l|rrrr}
      \hline\hline
     & \multicolumn{2}{c}{SVMfs} & \multicolumn{2}{c}{$best\_f$}\\
& Acc. $\mu$ &	Acc. $\sigma$ &  Acc. $\mu$ &	Acc. $\sigma$\\
ADJ & 0.829 & 0.019 & 0.821 & 0.016 \\
NOUN & 0.784 & 0.021 & 0.765 & 0.023 \\
VERBS & 0.782 & 0.052 & 0.744 & 0.046 \\
 \hline
		\end{tabular}
		}
	
	\setlength{\belowcaptionskip}{-0.1cm}
	\caption{Accuracy results for PoS using $SWN_3$}
	\label{tab:GI_POS}
\end{table}

Finally we test against the male and female ratings provided
by ANEW. As can be seen from Table \ref{tab:MF_general}, SWN
approaches are far more precise in predicting Male judgments rather
than Female ones (MAE$\mu$ goes from 0.392 to 0.323 with the best
formula and from 0.369 to 0.292 with $SVMfs$, both
differences are significant $p<0.001$).
Instead, in Table \ref{tab:MF_cont} -- which displays the results along gender
and polarity dimensions -- there is no statistically significant
difference in $MAE$ on positive words between male and female, while
there is a strong statistical significance for negative words ($p<0.001$).

Interestingly, there is also a large difference between positive and negative affective words (both for male and female dimensions). This difference is maximum for male scores on positive words compared to female scores on negative words (0.283 vs. 0.399, $p<0.001$). Recent work by Warriner et al. \shortcite{warriner2013norms} inspected the differences in prior polarity assessment due to gender. 



At this stage we can only note that prior polarities calculated with
SWN are closer to ANEW male annotations than female
ones. Understanding why this happens would require an accurate
examination of the methods used to create WordNet and SWN (which will be the focus of our future work).

\begin{table} [h]

	\centering
	{\footnotesize
		\begin{tabular}{l|rr|rr}
      \hline\hline
     & \multicolumn{2}{c|}{Male} & \multicolumn{2}{c}{female}\\
& MAE $\mu$ & MAE $\sigma$ &  MAE $\mu$ & MAE $\sigma$\\
\hline
SVMfs & 0.292 & 0.020 & 0.369 & 0.008 \\
best\_f & 0.323 & 0.022 & 0.392 & 0.010 \\
 \hline
		\end{tabular}
		}
	
	\setlength{\belowcaptionskip}{-0.1cm}
	\caption{MAE results for Male vs Female using $SWN_3$}
	\label{tab:MF_general}
\end{table}

\begin{table} [h]

	\centering
	{\footnotesize
		\begin{tabular}{l|rr|rr}
      \hline\hline
     & \multicolumn{2}{c|}{Male} & \multicolumn{2}{c}{female}  \\
& MAE $\mu$ & MAE $\sigma$ &  MAE $\mu$ & MAE $\sigma$ \\
\hline
Pos & 0.283 & 0.022 & 0.340 & 0.009\\ 
Neg & 0.301 & 0.029 & 0.399 & 0.013\\ 
\hline
		\end{tabular}
		}
	
	\setlength{\belowcaptionskip}{-0.1cm}
	\caption{MAE for Male/Female - Pos/Neg using $SWN_3$}
	\label{tab:MF_cont}
\end{table}

\section{Conclusions}
\label{sec:conclusions}

We have presented a study on the posterior-to-prior polarity issue, i.e. the problem of computing words' prior polarity starting from their posterior polarities. Using two different versions of SentiWordNet and 30 different approaches that have been proposed in the literature, we have shown that researchers have not paid sufficient attention to this issue. Indeed, we showed that the better variants outperform the others on different datasets both in regression and classification tasks, and that they can represent a fairer state-of-art baseline approach using SentiWordNet. On top of this, we also showed that these state-of-the-art formulae can be further outperformed using a learning framework that combines the various formulae together. We conclude our analysis with some experiments investigating the impact of word PoS and annotator gender in gold standards, showing interesting phenomena that requires further investigation.

\section*{Acknowledgments}
The authors thank Jos\'{e} Camargo De Souza for his help with feature selection. 
This work 
has been partially supported by the Trento RISE PerTe project.

\bibliographystyle{naaclhlt2013}
\bibliography{persuasive}

\begin{thebibliography}{}

\bibitem[\protect\citename{Agirre \bgroup et al.\egroup
  }2010]{agirre2010semeval}
E.~Agirre, O.L. De~Lacalle, C.~Fellbaum, S.K. Hsieh, M.~Tesconi, M.~Monachini,
  P.~Vossen, and R.~Segers.
\newblock 2010.
\newblock Semeval-2010 task 17: All-words word sense disambiguation on a
  specific domain.
\newblock In {\em Proceedings of the 5th International Workshop on Semantic
  Evaluation (IWSE '10)}, pages 75--80, Uppsala, Sweden.

\bibitem[\protect\citename{Agrawal and Siddiqui}2009]{agrawal2009using}
S.~Agrawal and T.J. Siddiqui.
\newblock 2009.
\newblock Using syntactic and contextual information for sentiment polarity
  analysis.
\newblock In {\em Proceedings of the 2nd International Conference on
  Interaction Sciences: Information Technology, Culture and Human (ICIS '09)},
  pages 620--623, Seoul, Republic of Korea.

\bibitem[\protect\citename{Baccianella \bgroup et al.\egroup
  }2010]{baccianella2010sentiwordnet}
S.~Baccianella, A.~Esuli, and F.~Sebastiani.
\newblock 2010.
\newblock {SentiWordNet} 3.0: An enhanced lexical resource for sentiment
  analysis and opinion mining.
\newblock In {\em Proceedings of the 7th Conference on International Language
  Resources and Evaluation (LREC '10)}, pages 2200--2204, Valletta, Malta.

\bibitem[\protect\citename{Bradley and Lang}1999]{bradley1999affective}
M.M. Bradley and P.J. Lang.
\newblock 1999.
\newblock Affective norms for {English} words ({ANEW}): Instruction manual and
  affective ratings.
\newblock Technical Report C-1, University of Florida.

\bibitem[\protect\citename{Chang and Lin}2011]{CC01a}
C.C. Chang and C.J. Lin.
\newblock 2011.
\newblock {LIBSVM}: A library for support vector machines.
\newblock {\em ACM Transactions on Intelligent Systems and Technology},
  2:27:1--27:27.

\bibitem[\protect\citename{Chaumartin}2007]{chaumartin2007upar7}
F.R. Chaumartin.
\newblock 2007.
\newblock {UPAR7}: A knowledge-based system for headline sentiment tagging.
\newblock In {\em Proceedings of the 4th International Workshop on Semantic
  Evaluations (IWSE '07)}, pages 422--425, Prague, Czech Republic.

\bibitem[\protect\citename{Chowdhury \bgroup et al.\egroup
  }2013]{chowdhury-EtAl:2013:SemEval-2013}
F.M. Chowdhury, M.~Guerini, S.~Tonelli, and A.~Lavelli.
\newblock 2013.
\newblock Fbk: Sentiment analysis in twitter with tweetsted.
\newblock In {\em Second Joint Conference on Lexical and Computational
  Semantics (*SEM): Proceedings of the Seventh International Workshop on
  Semantic Evaluation (SemEval '13)}, volume~2, pages 466--470, Atlanta,
  Georgia, USA, June.

\bibitem[\protect\citename{Cohn and Specia}2013]{cohnmodelling}
T.~Cohn and L.~Specia.
\newblock 2013.
\newblock Modelling annotator bias with multi-task gaussian processes: An
  application to machine translation quality estimation.
\newblock In {\em Proceedings of the 51th Annual Meeting of the Association for
  Computational Linguistics (ACL '13)}, pages 32--42, Sofia, Bulgaria.

\bibitem[\protect\citename{Denecke}2008]{denecke2008accessing}
K.~Denecke.
\newblock 2008.
\newblock Accessing medical experiences and information.
\newblock In {\em Proceedings of the 18th European Conference on Artificial
  Intelligence, Workshop on Mining Social Data (MSoDa '08)}, Patras, Greece.

\bibitem[\protect\citename{Denecke}2009]{denecke2009sentiwordnet}
K.~Denecke.
\newblock 2009.
\newblock Are {SentiWordNet} scores suited for multi-domain sentiment
  classification?
\newblock In {\em Proceedings of the 4th International Conference on Digital
  Information Management (ICDIM '09)}, pages 32--37, Ann Arbor, MI, USA.

\bibitem[\protect\citename{Devitt and Ahmad}2007]{devitt2007sentiment}
A.~Devitt and K.~Ahmad.
\newblock 2007.
\newblock Sentiment polarity identification in financial news: A cohesion-based
  approach.
\newblock In {\em Proceedings of the 45th Annual Meeting of the Association for
  Computational Linguistics (ACL '07)}, pages 984--991, Prague, Czech Republic.

\bibitem[\protect\citename{Dunphy \bgroup et al.\egroup }1974]{harvardiv}
D.C. Dunphy, C.G. Bullard, and E.E.M. Crossing.
\newblock 1974.
\newblock Validation of the {General Inquirer Harvard IV Dictionary}.
\newblock Paper presented at the Pisa Conference on Content Analysis.

\bibitem[\protect\citename{Esuli and Sebastiani}2006]{Esuli06}
A.~Esuli and F.~Sebastiani.
\newblock 2006.
\newblock {SentiWordNet}: A publicly available lexical resource for opinion
  mining.
\newblock In {\em Proceedings of the 5th Conference on International Language
  Resources and Evaluation (LREC '06)}, pages 417--422, Genova, Italy.

\bibitem[\protect\citename{Gatti and Guerini}2012]{gatti-guerini:2012:POSTERS}
L.~Gatti and M.~Guerini.
\newblock 2012.
\newblock Assessing sentiment strength in words prior polarities.
\newblock In {\em Proceedings of the 24th International Conference on
  Computational Linguistics (COLING '12)}, pages 361--370, Mumbai, India.

\bibitem[\protect\citename{Giora}1997]{giora1997understanding}
R.~Giora.
\newblock 1997.
\newblock Understanding figurative and literal language: The graded salience
  hypothesis.
\newblock {\em Cognitive Linguistics}, 8:183--206.

\bibitem[\protect\citename{Guerini \bgroup et al.\egroup }2008]{Guerini2008}
M.~Guerini, O.~Stock, and C.~Strapparava.
\newblock 2008.
\newblock Valentino: A tool for valence shifting of natural language texts.
\newblock In {\em Proceedings of the 6th International Conference on Language
  Resources and Evaluation (LREC '08)}, pages 243--246, Marrakech, Morocco.

\bibitem[\protect\citename{Lasswell and Namenwirth}1969]{lasswell1969lasswell}
H.D. Lasswell and J.Z. Namenwirth.
\newblock 1969.
\newblock The {Lasswell} value dictionary.
\newblock {\em New Haven}.

\bibitem[\protect\citename{Liu and Zhang}2012]{liu2012survey}
B.~Liu and L.~Zhang.
\newblock 2012.
\newblock A survey of opinion mining and sentiment analysis.
\newblock {\em Mining Text Data}, pages 415--463.

\bibitem[\protect\citename{Meinshausen and B\"{u}hlmann}2010]{Meinshausen2010}
N.~Meinshausen and P.~B\"{u}hlmann.
\newblock 2010.
\newblock {Stability selection}.
\newblock {\em Journal of the Royal Statistical Society: Series B (Statistical
  Methodology)}, 72(4):417--473.

\bibitem[\protect\citename{Neviarouskaya \bgroup et al.\egroup
  }2009]{neviarouskaya2009sentiful}
A.~Neviarouskaya, H.~Prendinger, and M.~Ishizuka.
\newblock 2009.
\newblock Sentiful: Generating a reliable lexicon for sentiment analysis.
\newblock In {\em Proceedings of the 3rd Affective Computing and Intelligent
  Interaction (ACII '09)}, pages 363---368, Amsterdam, Netherlands.

\bibitem[\protect\citename{Neviarouskaya \bgroup et al.\egroup
  }2011]{neviarouskaya2011affect}
A.~Neviarouskaya, H.~Prendinger, and M.~Ishizuka.
\newblock 2011.
\newblock Affect analysis model: novel rule-based approach to affect sensing
  from text.
\newblock {\em Natural Language Engineering}, 17(1):95.

\bibitem[\protect\citename{\"Ozbal and Strapparava}2012]{ozbalcomputational}
G.~\"Ozbal and C.~Strapparava.
\newblock 2012.
\newblock A computational approach to the automation of creative naming.
\newblock In {\em Proceedings of the 50th Annual Meeting of the Association for
  Computational Linguistics (ACL '12)}, pages 703--711, Jeju Island, Korea.

\bibitem[\protect\citename{\"Ozbal \bgroup et al.\egroup }2012]{ozbal2012brand}
G.~\"Ozbal, C.~Strapparava, and M.~Guerini.
\newblock 2012.
\newblock Brand {Pitt}: A corpus to explore the art of naming.
\newblock In {\em Proceedings of the 8th International Conference on Language
  Resources and Evaluation (LREC '12)}, pages 1822--1828, Istanbul, Turkey.

\bibitem[\protect\citename{Paltoglou \bgroup et al.\egroup
  }2010]{Paltoglou2010}
G.~Paltoglou, M.~Thelwall, and K.~Buckley.
\newblock 2010.
\newblock Online textual communications annotated with grades of emotion
  strength.
\newblock In {\em Proceedings of the 3rd International Workshop of Emotion:
  Corpora for research on Emotion and Affect (satellite of LREC '10)}, pages
  25--31, Valletta, Malta.

\bibitem[\protect\citename{Pang and Lee}2008]{pang2008opinion}
B.~Pang and L.~Lee.
\newblock 2008.
\newblock Opinion mining and sentiment analysis.
\newblock {\em Foundations and Trends in Information Retrieval}, 2(1-2):1--135.

\bibitem[\protect\citename{Pianta \bgroup et al.\egroup
  }2008]{pianta2008textpro}
E.~Pianta, C.~Girardi, and R.~Zanoli.
\newblock 2008.
\newblock The {TextPro} tool suite.
\newblock In {\em Proceedings of the 6th International Conference on Language
  Resources and Evaluation (LREC '08)}, pages 2603--2607, Marrakech, Morocco.

\bibitem[\protect\citename{Piller}2003]{piller200310}
I.~Piller.
\newblock 2003.
\newblock Advertising as a site of language contact.
\newblock {\em Annual Review of Applied Linguistics}, 23:170--183.

\bibitem[\protect\citename{Polajnar \bgroup et al.\egroup
  }2011]{polajnar2011protein}
T.~Polajnar, S.~Rogers, and M.~Girolami.
\newblock 2011.
\newblock Protein interaction detection in sentences via gaussian processes: a
  preliminary evaluation.
\newblock {\em International journal of data mining and bioinformatics},
  5(1):52--72.

\bibitem[\protect\citename{Qu \bgroup et al.\egroup }2008]{qu2008sentence}
L.~Qu, C.~Toprak, N.~Jakob, and I.~Gurevych.
\newblock 2008.
\newblock Sentence level subjectivity and sentiment analysis experiments in
  {NTCIR-7 MOAT} challenge.
\newblock In {\em Proceedings of the 7th NTCIR Workshop Meeting (NTCIR '08)},
  pages 210--217, Tokyo, Japan.

\bibitem[\protect\citename{Rasmussen and Williams}2006]{2006gaussian}
C.E. Rasmussen and C.K.I. Williams.
\newblock 2006.
\newblock {\em Gaussian processes for machine learning}.
\newblock MIT Press.

\bibitem[\protect\citename{Shawe-Taylor and Cristianini}2004]{shawe2004kernel}
J.~Shawe-Taylor and N.~Cristianini.
\newblock 2004.
\newblock {\em Kernel methods for pattern analysis}.
\newblock Cambridge university press.

\bibitem[\protect\citename{Sing \bgroup et al.\egroup
  }2012]{sing2012development}
J.K. Sing, S.~Sarkar, and T.K. Mitra.
\newblock 2012.
\newblock Development of a novel algorithm for sentiment analysis based on
  adverb-adjective-noun combinations.
\newblock In {\em Proceedings of the 3rd National Conference on Emerging Trends
  and Applications in Computer Science (NCETACS '12)}, pages 38--40, Shillong,
  India.

\bibitem[\protect\citename{Stone \bgroup et al.\egroup }1966]{stone1966general}
P.J. Stone, D.C. Dunphy, and M.S. Smith.
\newblock 1966.
\newblock {\em The General Inquirer: A Computer Approach to Content Analysis.}
\newblock MIT press.

\bibitem[\protect\citename{Strapparava and Valitutti}2004]{strappaLREC04}
C.~Strapparava and A.~Valitutti.
\newblock 2004.
\newblock {W}ord{N}et-{A}ffect: an affective extension of {W}ord{N}et.
\newblock In {\em Proceedings of the 4th International Conference on Language
  Resources and Evaluation (LREC '04)}, pages 1083 -- 1086, Lisbon, Portugal.

\bibitem[\protect\citename{Taboada \bgroup et al.\egroup
  }2011]{taboada2011lexicon}
M.~Taboada, J.~Brooke, M.~Tofiloski, K.~Voll, and M.~Stede.
\newblock 2011.
\newblock Lexicon-based methods for sentiment analysis.
\newblock {\em Computational linguistics}, 37(2):267--307.

\bibitem[\protect\citename{Thet \bgroup et al.\egroup }2009]{thet2009sentiment}
T.T. Thet, J.C. Na, C.S.G. Khoo, and S.~Shakthikumar.
\newblock 2009.
\newblock Sentiment analysis of movie reviews on discussion boards using a
  linguistic approach.
\newblock In {\em Proceedings of the 1st international CIKM workshop on
  Topic-sentiment analysis for mass opinion (TSA '09)}, pages 81--84, Hong
  Kong.

\bibitem[\protect\citename{Wang and Manning}2012]{wangbaselines}
S.~Wang and C.D. Manning.
\newblock 2012.
\newblock Baselines and bigrams: Simple, good sentiment and topic
  classification.
\newblock In {\em Proceedings of the 50th Annual Meeting of the Association for
  Computational Linguistics (ACL '12)}, pages 90--94, Jeju Island, Korea.

\bibitem[\protect\citename{Warriner \bgroup et al.\egroup
  }2013]{warriner2013norms}
A.B. Warriner, V.~Kuperman, and M.~Brysbaert.
\newblock 2013.
\newblock Norms of valence, arousal, and dominance for 13,915 english lemmas.
\newblock {\em Behavior research methods}, pages 1--17.

\bibitem[\protect\citename{Weston \bgroup et al.\egroup
  }2000]{weston2000feature}
J.~Weston, S.~Mukherjee, O.~Chapelle, M.~Pontil, T.~Poggio, and V.~Vapnik.
\newblock 2000.
\newblock Feature selection for {SVMs}.
\newblock In {\em Proceedings of the 14th Conference on Neural Information
  Processing Systems (NIPS '00)}, pages 668--674, Denver, CO, USA.

\bibitem[\protect\citename{Williams and Barber}1998]{williams1998bayesian}
C.K.I. Williams and D.~Barber.
\newblock 1998.
\newblock Bayesian classification with gaussian processes.
\newblock {\em Pattern Analysis and Machine Intelligence, IEEE Transactions
  on}, 20(12):1342--1351.

\bibitem[\protect\citename{Wilson \bgroup et al.\egroup }2004]{wilson:AAAI-04}
T.~Wilson, J.~Wiebe, and R.~Hwa.
\newblock 2004.
\newblock Just how mad are you? {F}inding strong and weak opinion clauses.
\newblock In {\em Proceedings of the 19th National Conference on Artificial
  Intelligence (AAAI '04)}, pages 761--769, San Jose, CA, USA.

\bibitem[\protect\citename{Wilson \bgroup et al.\egroup
  }2005]{wilson2005recognizing}
T.~Wilson, J.~Wiebe, and P.~Hoffmann.
\newblock 2005.
\newblock Recognizing contextual polarity in phrase-level sentiment analysis.
\newblock In {\em Proceedings of the Conference on Human Language Technology
  and Empirical Methods in Natural Language Processing (HLT/EMNLP '05)}, pages
  347--354, Vancouver, Canada.

\bibitem[\protect\citename{Yeh}2000]{yeh2000more}
A.~Yeh.
\newblock 2000.
\newblock More accurate tests for the statistical significance of result
  differences.
\newblock In {\em Proceedings of the 18th International Conference on
  Computational Linguistics (COLING '00)}, pages 947--953, Saarbr\"{u}cken,
  Germany.

\end{thebibliography}

\end{document}